\documentclass[runningheads]{llncs}
\usepackage{graphicx}
\usepackage[pagebackref=true,breaklinks=true,colorlinks,bookmarks=true]{hyperref} 
\usepackage{tikz}
\usepackage{comment}
\usepackage{amsmath,amssymb} 
\usepackage{color}
\usepackage{url}
\usepackage{caption}
\usepackage[capitalize]{cleveref}

\usepackage[accsupp]{axessibility}  
\usepackage[symbol]{footmisc}

\def\etal{{\em et al.}}

\begin{document}

\pagestyle{headings}
\mainmatter
\def\ECCVSubNumber{4261}  

\title{Learning Degradation Representations for Image Deblurring} 


\titlerunning{Learning Degradation Representations for Image Deblurring}
%
\author{Dasong Li\textsuperscript{1} \and
Yi Zhang\textsuperscript{1} \and
Ka Chun Cheung\textsuperscript{2}  \and 
Xiaogang Wang\textsuperscript{1,4}  \and \\
Hongwei Qin\textsuperscript{3}\protect\footnotemark[2] \and
Hongsheng Li\inst{1,4,5}\protect\footnotemark[2]  \\
\protect \textsuperscript{1}MMLab, CUHK \quad
\protect \textsuperscript{2}NVIDIA AI Technology Center \quad
\protect \textsuperscript{3}SenseTime Research \\
\protect \textsuperscript{4}Centre for Perceptual and Interactive Intelligence Limited \ 
\protect \textsuperscript{5}Xidian University \\
{\tt\small \{dasongli@link, hsli@ee\}.cuhk.edu.hk, qinhongwei@sensetime.com}
}
\footnotetext[2]{Corresponding authors}
\authorrunning{Li et al.}
\institute{}
\maketitle

\begin{abstract}
In various learning-based image restoration tasks, such as image denoising and image super-resolution, the degradation representations were widely used to model the degradation process and handle complicated degradation patterns.
However, they are less explored in learning-based image deblurring as blur kernel estimation cannot perform well in real-world challenging cases. We argue that it is particularly necessary for image deblurring to model degradation representations since blurry patterns typically show much larger variations than noisy patterns or high-frequency textures.
In this paper, we propose a framework to learn spatially adaptive degradation representations of blurry images. A novel joint image reblurring and deblurring learning process is presented to improve the expressiveness of degradation representations. 
To make learned degradation representations effective in reblurring and deblurring, we propose a Multi-Scale Degradation Injection Network (MSDI-Net) to integrate them into the neural networks. With the integration, MSDI-Net can handle various and complicated blurry patterns adaptively. 
Experiments on the GoPro and RealBlur datasets demonstrate that our proposed deblurring framework with the learned degradation representations outperforms state-of-the-art methods with appealing improvements. The code is released at \url{https://github.com/dasongli1/Learning_degradation}.
\keywords{Image Deblurring, Degradation Representations}
\end{abstract}

\section{Introduction}
\label{sec:intro}
Image restoration is required to handle various and complicated degradation patterns produced in different degradation processes. The degradation representations act as a crucial component to model the degradation processes and handle complicated degradation patterns, such as different noise levels in image denoising \cite{zhang2018ffdnet,Guo2019Cbdnet,MildenhallKPN18} and different combinations of Gaussian blurs and motion blurs in blind super-resolution \cite{unsupervised_degradation}. 
However, the degradation representations are less exploited in learning-based deblurring methods and have not been well integrated into state-of-the-art deblurring networks.

The general blurring process can be formulated as
\begin{equation}
    y = F(x, k) + \eta,
\end{equation}
where $x$ and $y$ are sharp image and blurry image respectively.  $F(x,k)$ is usually modeled as a blurring operator with kernel $k$. $\eta$ represents the Gaussian noise.

A popular paradigm for image deblurring is based on the Maximum A Posterior (MAP) estimate framework,
\begin{equation}
    (k, x) = \arg \max \mathbb{P}(y|x,k)\mathbb{P}(x)\mathbb{P}(k),
\end{equation}
where $\mathbb{P}(x)$ and $\mathbb{P}(k)$ model the priors of the clean images and the blur kernels. 
Many handcrafted priors for modeling $\mathbb{P}(x)$ and $\mathbb{P}(k)$ have been proposed ~\cite{blind_dark_channel,blind_spectral,hyper_laplacian,total_variation}. But most of them are insufficient in characterizing the clean images and blur kernels accurately. 
Furthermore, the operator $F$ is generally modeled as a convolution operation in conventional MAP frameworks, which does not hold in practice and causes unpleasing artifacts on real-world challenging cases.

Based on the limitations of kernel-based blurring modeling, a series of kernel-free approaches \cite{gao2019dynamic,tao2018srndeblur,HINet,Zamir2021MPRNet,MIMO_UNet} are proposed to directly learn the mapping from blurry images to corresponding sharp images. While those methods outperform previous deblurring methods significantly, their performances are still limited in complicated blurry patterns, due to the lack of explicit modeling of the degradation process. This is because, unlike denoising methods, where the noise level might be similar across different images, blurring of different images generally have totally different patterns and cannot be well handled by fixed-weight networks without considering the degradation process.
To combine the modeling of degradation and learning-based deblurring, recent works~\cite{Exploring_blur-CVPR21,Blind_DIP} propose to learn explicit degradation representations by using Deep Image Prior (DIP) \cite{DIP_2018_CVPR} to reparameterize the kernel $k$ and the sharp image $x$.
This inevitably involves the time-consuming iterative inverse optimization and hyperparameter tuning of DIP to adapt to the deblurring process.
Moreover, degradation representations have not been taken as a common component in SOTA deblurring methods~\cite{HINet,Zamir2021MPRNet}.

In this paper, we propose to learn explicit degradation representations with a novel joint sharp-to-blurry image reblurring and blurry-to-sharp image deblurring learning framework. 
Specifically, the degradation representations are learned in the process of sharp-to-blurry image reblurring. The process takes as input a blurry image and learns the degradation representations as a multi-channel spatial latent map to encode the spatially varying blur patterns in replacement of the conventional convolutional blur kernels or the DIP prior. 
A reblurring generator then takes as input the latent degradation map and the original sharp image and reblurs the sharp image back to its corresponding blurry image.

To effectively integrate the learned degradation representations into the reblurring process, we introduce a multi-scale degradation injection network (MSDI-Net) for achieving conditional image reblurring. 
The network adopts a U-Net like architecture \cite{U-Net}. The sharp image is fed into the encoder of the U-Net but the
degradation is input into the encoder-decoder via the skip-connections to modulate the shortcut encoder feature maps. Specifically, the latent degradation map
is gradually upsampled via nearest-neighbor interpolation and a convolution layer to multiple resolutions and are then used to predict spatially varying weighting and bias parameters of the shortcut feature maps at each corresponding resolution.
In this way, the learning of the latent degradation representation is supervised by the original blurry image for sharp-to-blurry image reblurring.


To make the learned degradation representations contributing to image deblurring, another blurry-to-sharp image generator network is also introduced, which shares the same MSDI-Net architecture but does not share weights with the reblurring generator. The learned degradation representations are processed similarly to deblur a blurry input image. 
Specifically, its encoder takes the blurry image as input, while the latent degradation representations are used to modulate the blurry image's encoder-decoder shortcut feature maps at multiple resolutions. 
With the help of learned degradation map, the deblurring generator can handle complicated spatially varying blurry patterns, which are adaptively learned from the data to optimize both reblurring and deblurring tasks. 

The main contributions of this work are two-fold:
1) We propose a novel joint framework for learning both sharp-to-blurry image reblurring and blurry-to-sharp image deblurring to adaptively encode spatially varying degradations and model the image blurring process, which in turn, benefits the image deblurring performance. 
2) The proposed joint reblurring and deblurring framework outperforms state-of-the-art image deblurring methods on the widely used GoPro \cite{deblur-multi-scale} and RealBlur \cite{realblur} datasets.

\section{Related Work}
In this section, we briefly talk about the related works of image restoration with degradation representations and different image deblurring methods.

\noindent\textbf{Image restoration with degradation representations.}
Image restoration tasks are usually required to handle different and complicated degradations on real-world applications. 
The degradation representations have been exploited and taken as one crucial component in several image restoration tasks, such as image denoising and image super-resolution. In image denoising, \cite{zhang2018ffdnet,MildenhallKPN18,Guo2019Cbdnet} take the noise variance as one network input to adaptively handle various noise strengths. Several practical denoising methods stabilize the noise variance caused by various ISO \cite{PMRID} and the property of Poisson-Gaussian distribution \cite{GAT,Li2022Efficient}.
Similarly, image blind super-resolution is required to handle various degradations (different Gaussian blurs, motion blurs, and noises) on real-world applications. 
\cite{unsupervised_degradation} proposes an unsupervised learning scheme for learning degradation representations based on the assumption that the degradation is the same in an image but can vary for different images. However, this assumption does not hold in image deblurring. 

\noindent\textbf{Optimization-based deblurring.}
A popular approach for image deblurring is based on the Maximum A Posterior (MAP). Most MAP-based methods focus on finding good priors for sharp images and blur kernels. 
Many priors are designed to model clean images and blur kernels. They include total variation (TV) \cite{total_variation}, hyper-Laplacian prior \cite{hyper_laplacian}, $l_0$-norm gradient prior \cite{L0_norm} and sparse image priors \cite{sparse_prior}.
They all assume the blur kernel is linear and uniform and can be represented as a convolution kernel. However, this assumption does not hold for real-world blurring with the non-uniform kernels. Some non-uniform deblurring methods \cite{non-uniform-deblurring,restore_spatially_variant,rotational_motion_deblurring,removing_non_uniform_motion} are proposed based on the assumption that the blur is locally uniform. They are not practical even with high computational costs.

\noindent\textbf{Learning-based deblurring.}
Many deep deblurring models have been proposed over the past few years. 
Earlier attempts \cite{Learning_to_deblur,Learning_CNN_non_uniform} utilize deep convolution neural networks to facilitate blur kernel estimation. However, there are several limitations \cite{deblur-multi-scale} in estimating kernels. 1) Simple kernel convolution is not practical on real-world challenging cases. 2) The incorrect kernel estimations, caused by noise and large motions, may cause unpleasing artifacts. 3) Estimating spatially varying kernels requires a huge amount of computation. To avoid the above limitations, a series of kernel-free methods \cite{tao2018srndeblur,deblur-multi-scale,kupyn2018deblurgan,kupyn2019deblurgan,deepstacked,gao2019dynamic,suin2020spatially,park2020multi,zhang2020deblurring,Zamir2021MPRNet,MIMO_UNet,HINet} are proposed with much better performances. Nah et al. \cite{deblur-multi-scale} propose a multi-scale network for image deblurring. Similarly, Tao et al. \cite{tao2018srndeblur} propose a scale-recurrent structure for image deblurring. Adversarial training is also introduced in image deblurring \cite{kupyn2018deblurgan,kupyn2019deblurgan}. Chen \etal~\cite{reblur2deblur} introduces a reblur2deblur framework for video deblurring. Zhang \etal~\cite{zhang2020deblurring} proposes a reblurring network to synthesize additional blurry training images. The reblurring network and deblurring network are separate in both two methods. In this work, we combine reblurring network and deblurring network in learning the degradation representations.
Recently, multi-stage approaches \cite{deepstacked,suin2020spatially,Zamir2021MPRNet,HINet} achieve impressive performance against previous methods.
While those deblurring networks outperform the traditional deblurring methods significantly, their performances are still limited due to the lack of explicit degradation modeling. 

\noindent\textbf{Learning Blurring Degradation Representations.}
While the explicit degradation representations have shown convincing improvements on many low-level vision tasks, it is rarely explored in learning-based deblurring methods. SelfDeblur~\cite{Blind_DIP} introduces the Deep Image Prior (DIP) \cite{DIP_2018_CVPR} to model the clean images and the kernels separately. But, it assumes the blur kernels are linear and uniform, which is not practical in complicated real-world scenes. Tran \etal~\cite{Exploring_blur-CVPR21} address the limitation by introducing the explicit representation for the blur kernels and the blur operators and reparameterizing the degradation and the sharp image by using DIP. This inevitably involves alternative optimization, which is time-consuming.   
The learned representations cannot improve the performance of existing deblurring networks directly and their application is limited to time-consuming DIP-based optimization methods.

\begin{figure*}[t]
  \centering
   \includegraphics[width=1.0\linewidth]{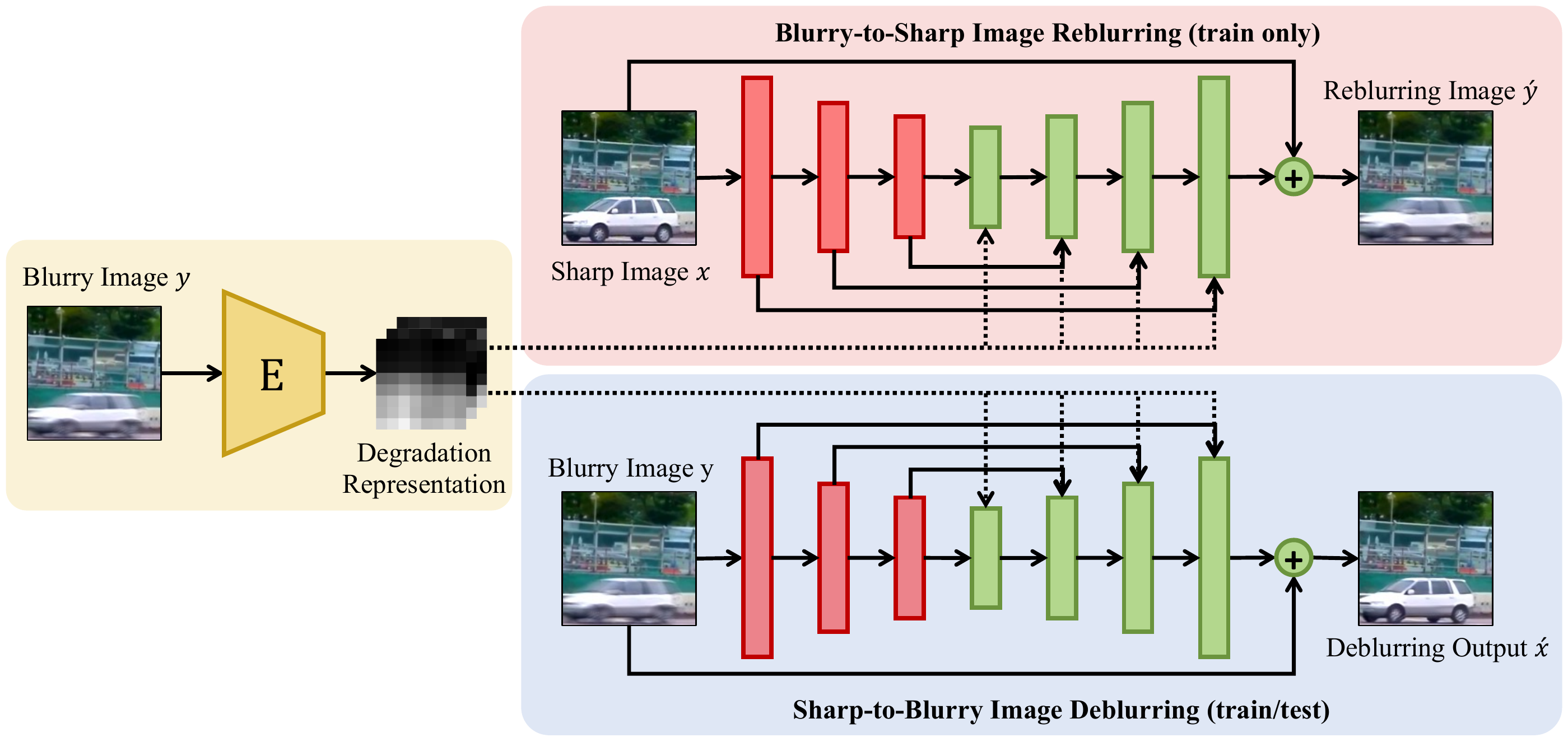}
   \caption{Learning degradation representations with reblurring and deblurring.}
   \label{fig:overview}
\end{figure*}

\section{Methodology}

In this section, we first introduce a joint learning framework for both sharp-to-blurry image reblurring and blurry-to-sharp image deblurring to encode latent spatially varying degradation representations from blurry images. A blur-aware loss is introduced to enhance the image deblurring performance.
To more effectively integrate the latent degradation representations for reblurring and deblurring, a multi-scale degradation injection network is proposed for both tasks.

\subsection{Learning Degradation Representations from Joint Reblurring and Deblurring} \label{sec31}
Most existing deblurring methods \cite{Learning_to_deblur,blind_dark_channel,blind_spectral,hyper_laplacian,total_variation} take the blur kernels as the degradation representations and model the blurring process as a convolution on the input image. However, the simple kernel convolution is not practical on real-world challenging cases and it is usually difficult to estimate the blur kernels in large motions and spatially varying blurring cases. 
We propose to encode latent degradation representations from blurry images via the joint learning of image reblurring and deblurring. 
As shown in Fig.~\ref{fig:overview}, we introduce an encoder $E$ to encode the blurry image $y$ into the degradation representations $E(y)$, which is modeled as a multi-channel latent map encoding 2D spatially varying blurring degradation in a latent space.
Then an image reblurring generator $G_r$ and an image deblurring generator $G_d$ are introduced to generate the reblurred image $\acute{y}$ and the deblurred image $\acute{x}$, respectively. 
The degradation representations do not only help the reblurring generator $G_r$ model the degradation process, but also help the deblurring network handle complicated spatially varying degradation patterns.
The joint training of reblurring and deblurring strengthens the expressiveness of learned degradation representations.

\noindent\textbf{Sharp-to-blurry image reblurring.} 
Different from modeling the blurring process as a convolution, our generator $G_r$ models the degradation process via learning to generate the blurry image $\acute{y}$, given the sharp image $x$ and the corresponding degradation representations $E(y)$. 
In addition, instead of generating the whole blurry image from scratch, the reblurring generator $G_r$ learns to predict the residual between the sharp image $x$ and the blurry image $y$. 
The learning of the blurring degradation process in our framework is therefore formulated as generating a blurry image $\acute{y}$ from its clean image $x$ as
\begin{equation}
     \acute{y} = x + G_r(x, E(y)),
\end{equation}
where $\acute{y}$ is the reblurred image conditioned on the sharp image $x$ and degradation representation $E(y)$.
With the residual learning, the encoder $E$ is encouraged to neglect the contents of the blurry image and to focus on disentangling the content-independent degradation representation $E(y)$ from the blurry image $y$.

Note that most existing encoder-decoder networks, such as VAE \cite{VAE}, can also learn implicit image representations via image reconstruction. They aim at encoding the whole-image contents into the latent representations. However, our framework aims at encoding only the degradation information by predicting the residuals between the sharp and blurry images. 
Such a task actually has a lower difficulty level than reconstructing all contents of an input image and therefore leads to encoding better blurring degradation representations.

We first tried the mainstream $L_1$ distance as the loss function. But the $L_1$ loss function merely measures the pixel-wise distance and cannot properly describe the similarity of the blurry patterns of two images. Then the reblurring generator $G_r$ cannot well generate the blurry images with $L_1$ loss, which harms the learning of degradation representation. 
Therefore, we further resort to perceptual loss \cite{Johnson2016Perceptual} and adversarial training \cite{GAN} to distinguish different degradation patterns in training. 
Adversarial loss is applied on the output of the generator $G_r$ to distinguish real and fake blurry images so that the decoder can well model the blurring process and improve the expressiveness of degradation representation.
We take the hinge loss \cite{lim2017geometric,SAGAN,SNGAN,park2019SPADE} as the adversarial loss to help the reblurring generator $G_r$ model the degradation (blurring) process and improve the expressiveness of learned degradation representations. We train the reblurring generator $G_r$ to generate the blurry images with the multi-scale discriminator $D$ used in \cite{wang2018pix2pixHD}. The training objective for image reblurring is formulated as 
\begin{equation}
\begin{aligned}
    L_G &= - E_{x\sim p_{\mathrm{data}}} D(x, \acute{y}) + \lambda_1 L_{\mathrm{perceptual}}(y,\acute{y}), \\
    L_D &= - E_{(x,y)\sim p_{\mathrm{data}}} [\mathrm{min}(0, -1 + D(x,y))] - E_{x\sim p_{\mathrm{data}}} [\mathrm{min} (0, -1-D(x, \acute{y}))],
\end{aligned}
\end{equation}
where $\lambda_1$ balances the $L_1$ loss and the discriminator loss for image reblurring, and the discriminator $D$ is a conditional discriminator conditioning on the sharp image $x$. 
Conditioning on the sharp image $x$, the conditional discriminator $D$ can focus on whether the reblurred image $\acute{y}$ has the same image contents with the corresponding sharp image $x$. 
The reblurring helps extract content-independent degradation information (shown in Fig), which is different with the content-dependent conditional networks \cite{wang2018sftgan,spatially_variant_recurrent,zhou2019stfan}.

\noindent\textbf{Blurry-to-sharp image deblurring.} To make the learned degradation representations contributing to image deblurring, we also model the image deblurring process with the image reblurring process jointly. 
The image deblurring generator $G_d$ follows a similar design to that of $G_r$. The deblurring generator $G_d$ is only required to learn the deblurring residuals between the blurry image $y$ and its corresponding sharp image $x$, and the learned degradation representations $E(y)$. The blurry-to-sharp image deblurring is modeled as
\begin{equation}
     \acute{x} = y + G_d(y, E(y)).
\end{equation}
Thanks to the 2D learnable degradation representations, the image deblurring generator $G_d$ is aware of the spatially varying blurry patterns and thus can adaptively handle various and complicated degradation patterns. The learning of image deblurring makes the learned degradation representations adapted to the deblurring task. The loss function for image deblurring is formulated as
\begin{equation}
    L_(x, \acute{x}) = \lambda_2 L_1(x, \acute{x}).
\end{equation}
\noindent\textbf{Discussion of the learned degradation representation.}
The learned degradation representation has two main advantages against conventional kernel modeling: 1) Our degradation representations can learn non-uniform spatially varying degradations effectively. 
\cref{fig:overview,fig:decoupling} show that the encoder can distinguishe different degradation representations. 
2) Interpolating on the latent space of representations can generate blurry images with controllable blurry levels (as shown in \cref{fig:interpolation}). The representations are also content-independent (as shown in \cref{fig:decoupling}), which is different from previous conditional networks \cite{wang2018sftgan,spatially_variant_recurrent,zhou2019stfan}. Built on this latent space, the representations show better interpretability and expressiveness.

\subsection{Image Deblurring with Learned Degradation Representations}
After obtaining the pre-trained encoder $E$, we freeze the pre-trained encoder and re-train the deblurring generator $G_d$ to illustrate that our improvement is not from the complicated framework but the learned degradation representations. To further demonstrate the generality of learned degradation representations, we also train the deblurring generator $G_d$ on the RealBlur dataset \cite{realblur} with the encoder trained on the GoPro dataset \cite{deblur-multi-scale}.

Following HINet \cite{HINet}, we select the Peak Signal-to-Noise Ratio (PSNR) loss as the main supervision. We also utilize a blur-aware loss function as extra supervision. The well-trained encoder $E$ should be quite sensitive to capture various and even subtle blur patterns. We, therefore, define the blur-aware loss as the distance between degradation encoder features of the ground-truth sharp image $x$ and the estimated sharp image $\acute{x}$. 
The deviation between the encoder feature maps $\|E(x)- E(\acute{x})\|_1$ of images $x$ and $\acute{x}$ can give more weights on the remaining blurry regions of the network output $\acute{x}$. Similar to perceptual loss \cite{Johnson2016Perceptual}, the $L_1$ distances between the encoder feature maps can be calculated at multiple scales.
\begin{figure*}[t]
  \centering
   \includegraphics[width=1.0\linewidth]{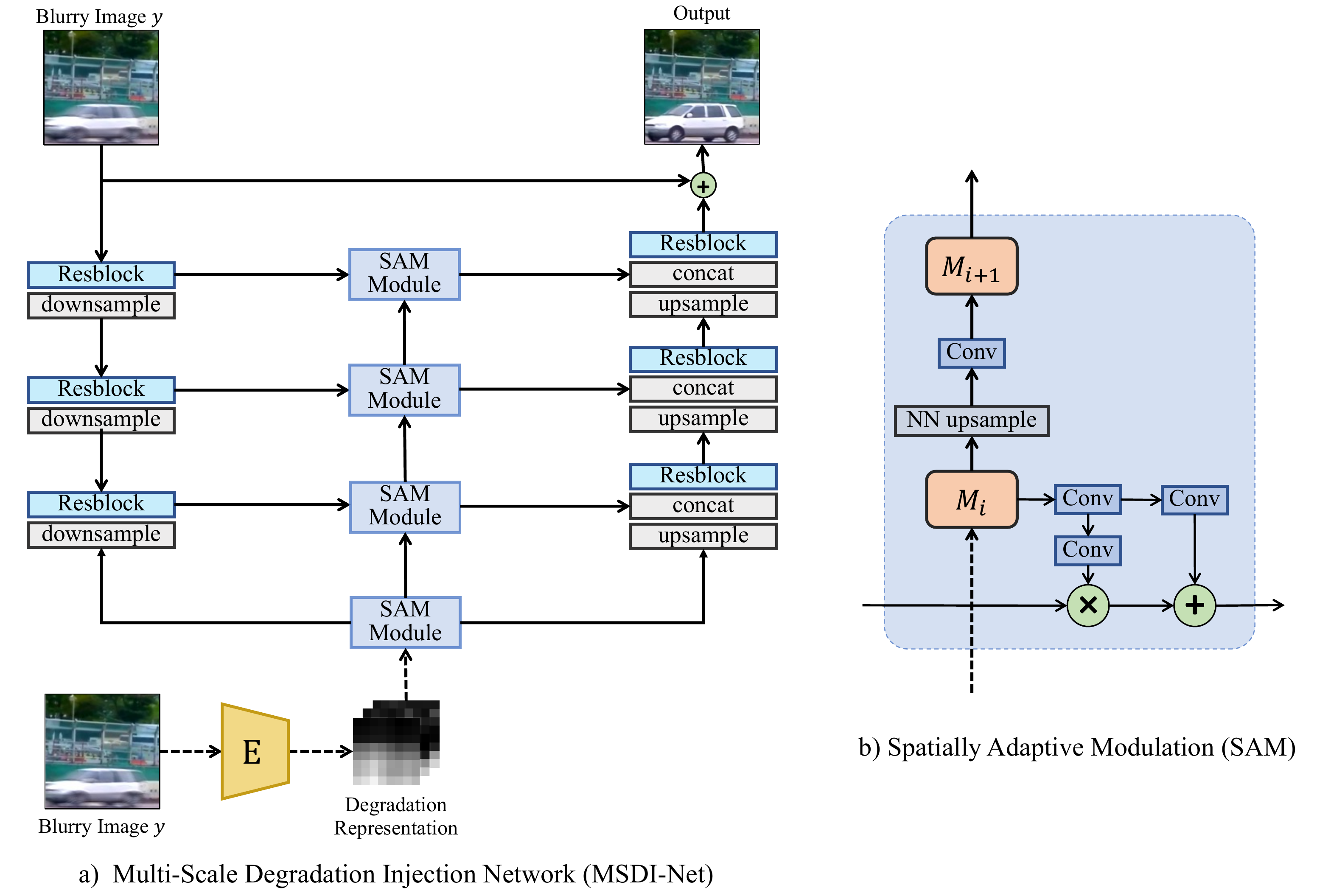}
   \caption{The network structure of multi-scale degradation injection network for image deblurring. Image reblurring shares the same structure.}
   \label{fig:deblur_structure}
\end{figure*}
The blur-aware loss $L_{\mathrm{blur}}$ is therefore formulated as
\begin{equation}
    \label{eq:blur-aware}
    L_{\mathrm{blur}}(\acute{x},x) = \sum_{i=1}^N \frac{1}{|E^{(i)}|} [||E^{(i)}(\acute{x}) - E^{(i)}(x)||_1],
\end{equation}
where $E^{(i)}$ denotes the $i$-th layer of the encoder $E$ and $|E^{(i)}|$ denotes the number of pixels in feature map $E^{(i)}(x)$. Using the blur-aware loss makes the deblurring generator $G_d$ pay more attention to the remaining blurry regions of the network output $\acute{x}$.
Then the deblurring objective is formulated as
\begin{equation}
    \label{eq:deblurring_loss}
    L(x,\acute{x})= \mathrm{PSNR}(x,\acute{x}) + \lambda_2 L_{\mathrm{blur}}(x,\acute{x}).
\end{equation}

\subsection{Multi-scale Degradation Injection Network}

To effectively integrate the degradation $E(y)$ to predict the reblurring and deblurring residuals, we propose a multi-scale degradation injection network (MSDI-Net) for both reblurring and deblurring.  Our reblurring generator $G_r$ and deblurring generator $G_d$ consists of two MSDI-Nets, which are stacked as HINet \cite{HINet} does. for simplicity, the overview of a MSDI-Net is shown in \cref{fig:deblur_structure}. 

The MSDI-Net consists of an encoder, a decoder, concatenation-based skip-connections, and a multi-scale degradation injection module. 
The first three modules of the MSDI-Net are widely adopted in U-Net like architectures \cite{U-Net}. The multi-scale degradation injection module modulates the feature of each skip-connection spatially, based on the learned degradation representation. The spatially variant modulation are explored in image synthesis \cite{park2019SPADE} and image super-resolution \cite{liang2021mutual}. We adopt it as the key to connecting learnable degradation representations to reblurring and deblurring. 
Let $f_i \in \mathbb{R}^{C_i \times H_i \times W_i}$ denote the features extracted at scale $i = 1, \dots, 5$ in the encoder. 
The extracted feature $f_i$ is passed to the decoder at the skip connection of scale $i$. 
Our MSDI-Net integrates the degradation representations into concatenation-based skip-connections of scale $i$. 
At scale $i$, we obtain the degradation map $M_i \in \mathbb{R}^{C_i \times H_i \times W_i}$ by a convolution-based upsampling block on the original degradation map $M_{i-1} \in \mathbb{R}^{C_{i-1} \times H_{i-1} \times W_{i-1}}$ ($H_{i} = 2 \times H_{i-1},W_{i} = 2 \times W_{i-1}$), which is implemented by a nearest-neighbor interpolation and a convolution layer to avoid checkerboard artifacts \cite{checkboardartifacts}. 
Then we utilize a spatially adaptive modulation (SAM) module to modulate the skip-connection feature $f_i$ at scale $i$. The spatially adaptive modulation modulates the feature map channels in a spatially varying manner with both predicted scaling and additions.
At the skip connection of scale $i$, we use several convolution ($3\times3$) layers on the degradation map $M_i$ to predict the modulating parameters $\gamma_i \in \mathbb{R}^{C_i \times H_i \times W_i}$ and $\beta_i \in \mathbb{R}^{C_i \times H_i \times W_i}$ respectively, which modulate the feature map $f_i$ as
\begin{equation}
    F_i = \gamma_i \odot f_i  + \beta_i,
\end{equation}
where $F_i$ is the modulated skip-connection features. Since the modulation parameters are predicted from the degradation representations, the learnable modulation of the feature channels makes the deblurring network aware of spatially varying degradations.
The degradation-aware feature $F_i$ is then concatenated with the decoder feature at scale $i$ and the last decoder layer predicts the image residuals for both image reblurring and deblurring.

Injecting the degradation representations enables the networks to handle various and complicated degradation patterns adaptively.
Injection at multiple scales improves the expressiveness of degradation representations by strengthening the connections between the degradation representations and two generators.

\section{Experiments}
\subsection{Dataset and implementation details}
We train and evaluate our method on the GoPro~\cite{deblur-multi-scale} and RealBlur datasets~\cite{realblur}. The GoPro dataset consists of 2,103 pairs of blurry and sharp images for training and 1,111 pairs for testing. The RealBlur dataset consists of 3,758 pairs for training and 980 pairs for testing. We first train the framework of reblurring and deblurring on GoPro dataset \cite{deblur-multi-scale}. 
We apply horizontal flipping and rotation as data augmentation and crop image patch of size $256 \times 256$ from the dataset for training.
$\lambda_1$ is set as 30 and $\lambda_2$ is set as 10. 
The networks of whole framework are trained with a batch size of 32 for 200k iterations. 
Then we freeze the weights of well-trained encoder $E$ and train the deblurring generator $G_d$ on the GoPro dataset \cite{deblur-multi-scale} and the RealBlur dataset \cite{realblur} respectively. $\lambda_3$ is set as 1. 
The deblurring generator $G_d$ is trained with a batch size of 64 for 400k iterations.
We use the Adam optimizer and the learning rate is set as $3 \times 10^{-4}$ at the beginning and decreased to $1 \times 10^{-7}$ following the cosine annealing strategy \cite{cosine_annealing}.

\begin{table}[t]
\centering

\begin{tabular}{l|ccc}
\hline

Method & Reference & PSNR & SSIM\\
\hline
DeblurGAN~\cite{kupyn2018deblurgan}& CVPR'18 & 28.70 & 0.858\\
DeblurGAN-v2~\cite{kupyn2019deblurgan} & ICCV'19 & 29.55 & 0.934\\
SRN~\cite{tao2018srndeblur} & CVPR'18 & 30.26 & 0.934\\
Gao et al.~\cite{gao2019dynamic} & CVPR'19 & 30.90 & 0.935\\
DBGAN~\cite{zhang2020deblurring} & CVPR'20 & 31.10 & 0.942\\
MT-RNN~\cite{park2020multi} & ECCV'20 & 31.15 & 0.945\\
DMPHN~\cite{deepstacked} & CVPR'19 & 31.20 & 0.940\\
Suin et al.~\cite{suin2020spatially} & CVPR'20 & 31.85 & 0.948\\
MIMO-UNet~\cite{MIMO_UNet} & ICCV'21 & 32.45 & 0.957 \\
MPRNet~\cite{Zamir2021MPRNet} & CVPR'21 & 32.66 & 0.959 \\
HINet~\cite{HINet} & CVPRW'21 & 32.71 & 0.959 \\
MPRNet-patch256~\cite{Zamir2021MPRNet} & CVPR'21 & \underline{32.96} & \underline{0.961} \\ 
\hline
Ours & ECCV'22 &\textbf{33.28} & \textbf{0.964}\\
\hline
\end{tabular}
\caption{Deblurring comparisons on GoPro~\cite{deblur-multi-scale} dataset. Best and second best scores are \textbf{highlighted} and \underline{underlined}.}
\label{tb:gopro}
\end{table}
\begin{table}[t]
\centering
		\setlength{\tabcolsep}{2pt}
		\begin{tabular}{l|c|c|c|c}
			\hline
			Model & Blurriest 10\% & Sharpest 10\% & All & MACs (G) \\  \hline 
			MPRNet-patch256 \cite{Zamir2021MPRNet} & 29.31 & 35.52 & 32.96 & 760.11 \\ \hline
			Ours & \textbf{29.65} & \textbf{35.58} & \textbf{33.28} & 336.43 \\ \hline
		\end{tabular}
	\caption{Detailed comparisons of MPRNet \cite{Zamir2021MPRNet} and our method on GoPro test dataset\cite{deblur-multi-scale}. MACs are estimated with the input size of 3$\times$256$\times$256. }
	\label{tb:macs}
\end{table}
\begin{figure*}[t]
  \centering
   \includegraphics[width=1.0\linewidth]{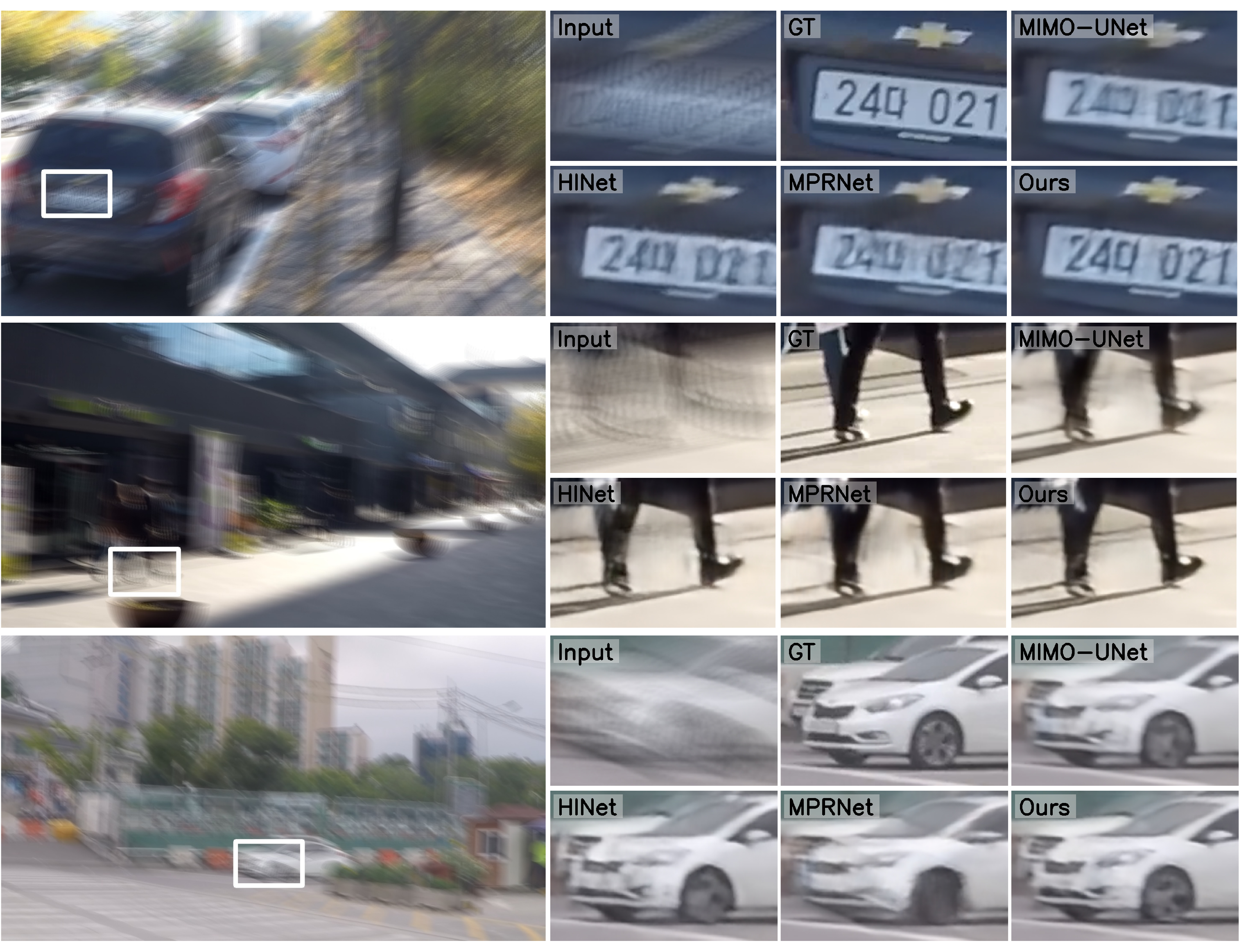}
   \caption{Visual comparisons for image deblurring on the GoPro test dataset \cite{deblur-multi-scale}. From left-top to right-bottom: blurry images, ground-truth images, and results obtained by MIMO-UNet \cite{MIMO_UNet}, HINet \cite{HINet}, MPRNet \cite{Zamir2021MPRNet} and our proposed method.}
   \label{fig:gopro_pics}
\end{figure*}

\subsection{Performance comparison}

We compare our method with state-of-the-art deblurring methods \cite{HINet,Zamir2021MPRNet,MIMO_UNet} on the GoPro test dataset \cite{deblur-multi-scale}. The quantitative results are reported in Table~\ref{tb:gopro}.
For testing, we slice the whole image into several $256\times 256$ patches and test all patches to report the results of HINet \cite{HINet}, MPRNet-patch256 \cite{Zamir2021MPRNet} and our method.
Our method achieves 0.45 dB improvement in terms of PSNR over the previous best-performing method HINet \cite{HINet}. 
\begin{figure*}[t]
  \centering
   \includegraphics[width=0.99\linewidth]{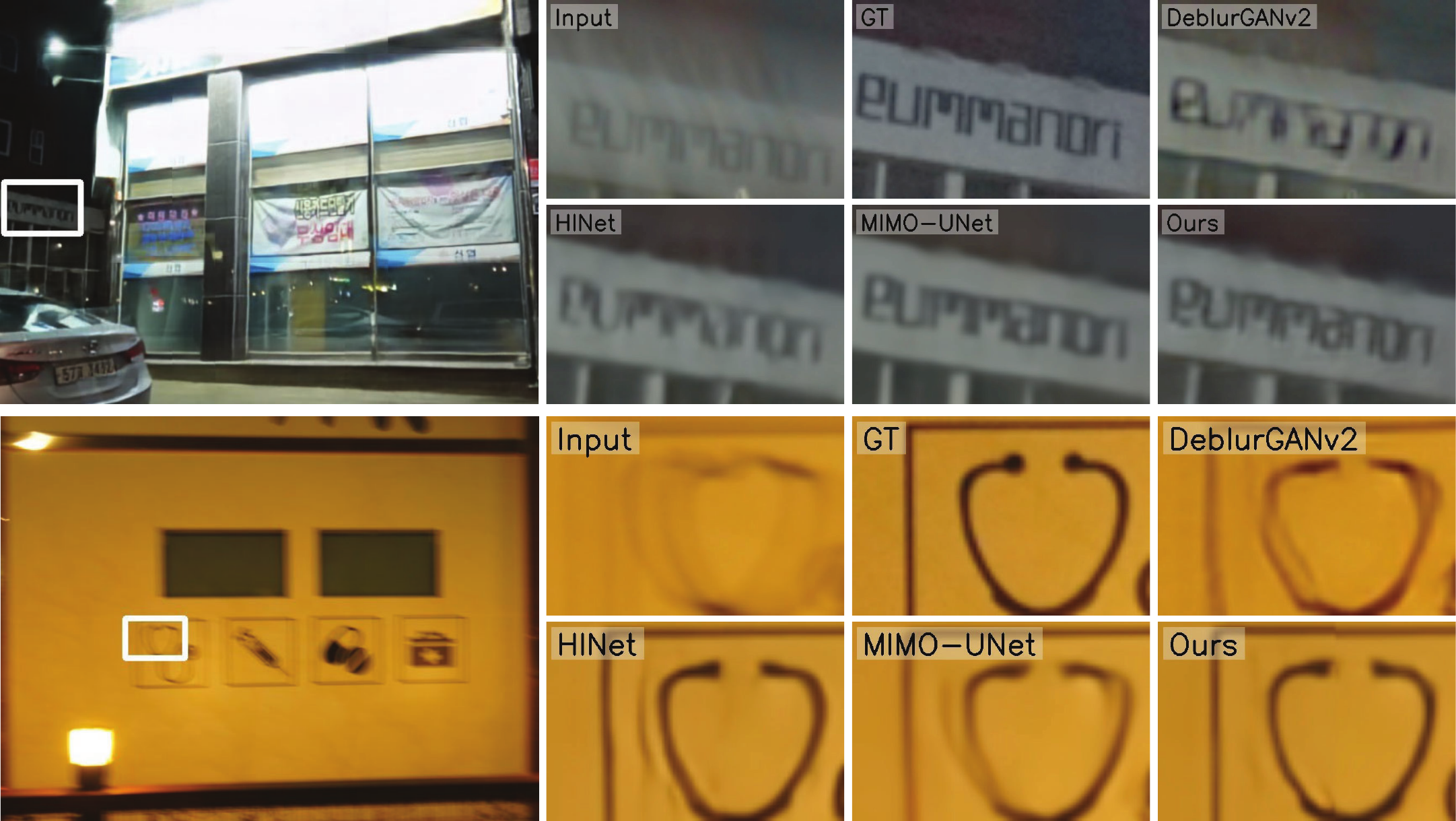}
   \caption{Visual comparisons for image deblurring on the RealBlur test dataset~\cite{realblur}. From left-top to right-bottom: blurry images, ground-truth images and resuls obtained by DeblurGANv2, HINet \cite{HINet}, MIMO-UNet \cite{MIMO_UNet}, our proposed method.}
   \label{fig:realblur}
\end{figure*}
\begin{table}[t]
\centering
		\begin{tabular}{l|ccc}
			\hline
			Method & Reference & PSNR & SSIM \\  \hline
			DeblurGAN-v2~\cite{kupyn2019deblurgan} & ICCV'19 & 29.69 & 0.870 \\
			SRN~\cite{tao2018srndeblur} & CVPR'18 & 31.38 & 0.909 \\ 
			MPRNet~\cite{Zamir2021MPRNet} & CVPR'21 & 31.76 & \underline{0.922} \\
			MIMO-UNet \cite{MIMO_UNet} & ICCV'21  & 32.05 & 0.921 \\ 
			HINet \cite{HINet} & CVPRW'21  & \underline{32.12} & 0.921 \\ \hline
			Ours & ECCV'22 & \textbf{32.35}  & \textbf{0.923} \\ \hline
		\end{tabular}
	\caption{Deblurring comparisons on the RealBlur test dataset~\cite{realblur}. }
	\label{tb:realblur}
\end{table}
To evaluate effectiveness and generality of learned degradation representations, we also evaluate our method on the RealBlur dataset. As listed in Table~\ref{tb:realblur}, our method achieves the best performance in terms of PSNR and SSIM. Since HINet \cite{HINet} does not release the model for RealBlur dataset \cite{realblur}, we train the HINet model on RealBlur dataset based on their released training code.
Note that we apply the degradation representations trained on GoPro dataset \cite{deblur-multi-scale} directly on RealBlur dataset \cite{realblur}.
Our method outperforms the previous SOTA HINet \cite{HINet} by 0.23 dB PSNR, which demonstrate the generality of learnable degradation representations.

In Table \ref{tb:macs}, we provide detailed comparisons between our method and MPRNet \cite{Zamir2021MPRNet}. 
We divide the whole gopro dataset into blurriest 10\% and sharpest 10\% as \cite{Son2021PVDNet} does. It is observed that the main improvement of our method is the improvement of 0.34dB PSNR in the blurriest 10\%, which demonstrates the advantage of proposed degradation learning. What's more, our method's computational cost is less than 50\% of MPRNet's \cite{Zamir2021MPRNet}.

Fig.~\ref{fig:gopro_pics} and Fig.~\ref{fig:realblur} show example deblurred results from the GoPro \cite{deblur-multi-scale} and RealBlur \cite{realblur} test sets by the evaluated approaches.
Our method produces sharper images and recovers more details in the regions of texts and moving objects, compared with other methods.

\subsection{Interpolation and decoupleness of degradation representations}

\begin{figure*}[ht]
    \small
    \begin{center}
    \begin{tabular}{ccccccc}
        \includegraphics[width=.15\linewidth]{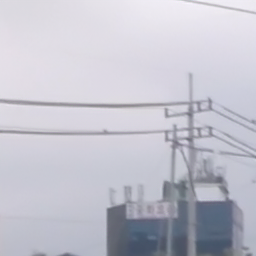} &
        \includegraphics[width=.15\linewidth]{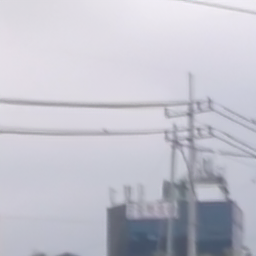} &
        \includegraphics[width=.15\linewidth]{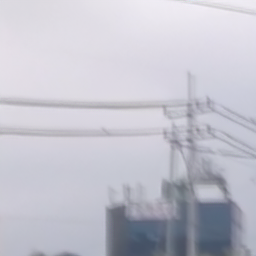} &
        \includegraphics[width=.15\linewidth]{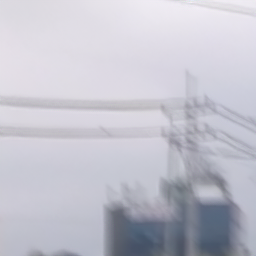} &
        \includegraphics[width=.15\linewidth]{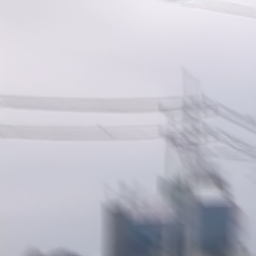} &
        \includegraphics[width=.15\linewidth]{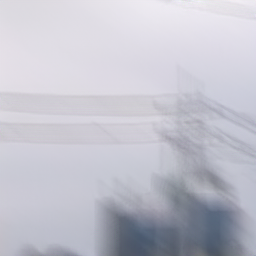} \\
        \includegraphics[width=.15\linewidth]{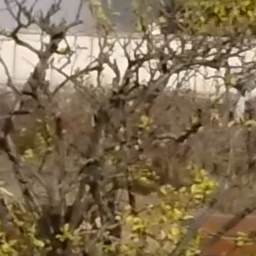} &
        \includegraphics[width=.15\linewidth]{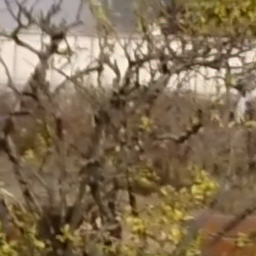} &
        \includegraphics[width=.15\linewidth]{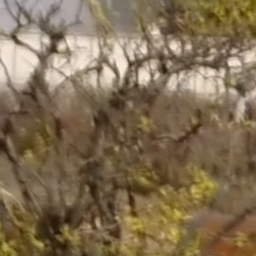} &
        \includegraphics[width=.15\linewidth]{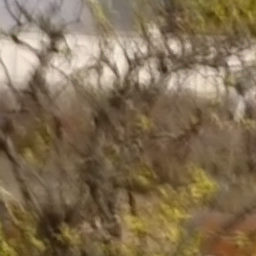} &
        \includegraphics[width=.15\linewidth]{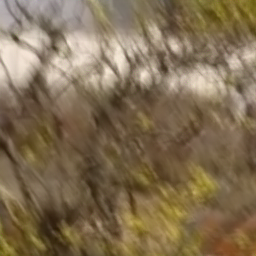} &
        \includegraphics[width=.15\linewidth]{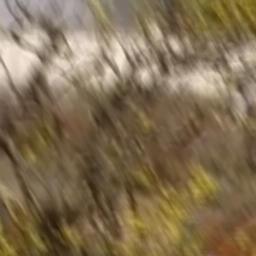} \\
    \end{tabular}
    \end{center}
    \vskip -0.1in
    \vspace{-3mm}
    \caption{Generating blurry images with linearly interpolated degradation representations. From left to right: the blurry level from sharp to blur.}
    \label{fig:interpolation}
    \vspace{-3mm}
\end{figure*}

To demonstrate the effectiveness of learned degradation representations, we study the interpolation and decoupleness of learned degradation representations. 
Given a blurry image $y$ and its corresponding sharp image $x$, we can obtain two degradation representations $E(y)$ and $E(x)$.  Then we obtain several intermediate degradation representations by interpolating from $E(y)$ to $E(x)$. The corresponding output of the decoder changes smoothly from sharp to blurry images. 
The blur interpolation in \cref{fig:interpolation} shows that our degradation representations are built on the latent space and accurately aware of different degradations.
We empirically validate the decoupleness of blur and image contents on GoPro dataset. We divide the GoPro test dataset into 555 pairs of images. For each pair of sharp images $\{A,B\}$, the corresponding blurry images are $\{\mathrm{blur}(A),\mathrm{blur}(B)\}$ and the degradation representations are obtained as $\{{\rm deg}_A, {\rm deg}_B\}$. The sharp image $A$ can be reblurred according to ${\rm deg}_B$ to obtain $\mathrm{ReBlur}(A, {\rm deg}_B)$. 
We use the average contextual similarity $\mathrm{CX}$ \cite{contextual} to measure pairwise image similarity. We have $\mathrm{CX}(\mathrm{blur}(A),A)= 2.72$, $\mathrm{CX}(\mathrm{ReBlur}(A,{\rm deg}_B),A)=2.65$, $\mathrm{CX}(\mathrm{blur}(A),B)=5.43$, and $\mathrm{CX}(\mathrm{ReBlur}(A,{\rm deg}_B),B)=5.39$, averaging over all the pairs. The similarities empirically show that  $\mathrm{ReBlur}(A, {\rm deg}_B)$ has similar contents with $A$ (the former two equations) but doesn't have similar contents with $B$ (the latter two equations). The visual examples on degradation decoupleness are provided in Figure~\ref{fig:decoupling}. The decoupleness of learned degradation representations shows our difference with the content-dependent conditional networks \cite{wang2018sftgan,spatially_variant_recurrent,zhou2019stfan} and the capacity of being a general operator to replace conventional blurring process.

\setlength{\tabcolsep}{2pt}
\begin{figure}[!t]
    \scriptsize
    \centering
    \setlength{\tabcolsep}{2pt}
    \begin{tabular}{@{} c c c c @{}}
        \includegraphics[width=0.18\linewidth]{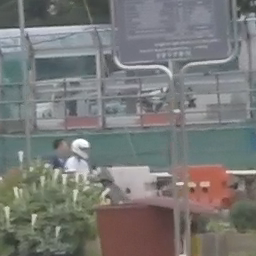} &
        \includegraphics[width=0.18\linewidth]{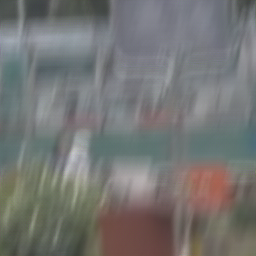}
        & \includegraphics[width=0.18\linewidth]{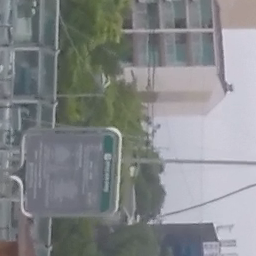} &
        \includegraphics[width=0.18\linewidth]{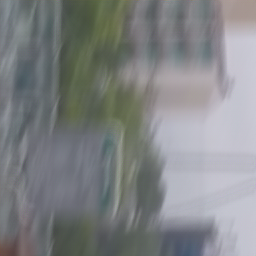}
        \\
        Sharp $B$ & $\mathrm{Blur}(B)$  & Sharp $A_1$ & ReBlur$(A_1)$ \\
        \includegraphics[width=0.18\linewidth]{figs/decoupling/input_1_1.png} &
        \includegraphics[width=0.18\linewidth]{figs/decoupling/gt_1_1.png} &
        \includegraphics[width=0.18\linewidth]{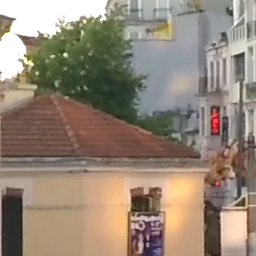} &
        \includegraphics[width=0.18\linewidth]{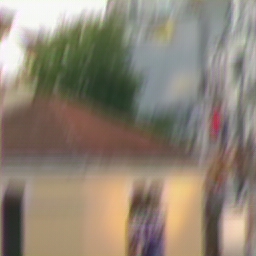}
        \\
        Sharp $A_2$  & ReBlur$(A_2)$ & Sharp $A_3$  & ReBlur$(A_3)$ \\
    \end{tabular}
    \caption{Reblurring $A_1,A_2,A_3$ with the degradation representation of $\mathrm{Blur}(B)$. }
  \label{fig:decoupling}
\end{figure}

\subsection{Ablation study}

\begin{table}[t]
\centering
		\begin{tabular}{l|cc}
			\hline
			Model & PSNR  \\  \hline 
			Ours w/o degradation & 32.81 \\
			Ours w/o reblurring & 33.09 \\ \hline
			Ours & \textbf{33.28}  \\\hline
			Ours w/o blur loss &  33.21 \\
			Our injection w/o multi-scale & 32.80 \\ 
			Our input w/ concat & 32.98 \\
			Ours w/ concat injection & 33.05  \\ \hline
			Ours & \textbf{33.28}  \\ \hline
		\end{tabular}
	\caption{The ablation study of image deblurring on GoPro test dataset\cite{deblur-multi-scale}.}
	\label{tb:ab_deblurring}
\end{table}
We evaluate the effectiveness of learning degradation representations and the multi-scale degradation injection network by revising one of the components of our model at a time. Table~\ref{tb:ab_deblurring} lists the performances of different settings on the GoPro test set \cite{deblur-multi-scale}.
We first remove the degradation encoder for learning the degradation representations and make $G_d$ only takes as input the blurry image (denoted as ``Ours w/o degradation'').
The performance suffers a large drop of 0.47dB PSNR. Then we remove the generator of image reblurring and reserve the encoder to provide additional encoding from the blurry images (denoted as ``Our w/o reblurring''), this operation causes a drop of 0.19 dB PSNR, demonstrating that reblurring indeed contributes to the learning of better degradation representations.
Then we remove the blur-aware loss function (Eq. (\ref{eq:blur-aware})). The training objective for deblurring becomes only the PSNR loss (denoted as ``Ours w/o blur loss''). The performance drops slightly by about 0.07 PSNR.
We further experiment with different ways of integrating degradation representations into the generators. 
When we remove the injection at multiple scales, the degradation representation is integrated just at the lowest resolution skip connection (denoted as ``Our injection w/o multi-scale''). The performance suffers a significant drop of 0.48 PSNR, which means that injecting the degradation representation into a single scale would affect the deblurring performance significantly.
Then we replace the modulation and test on concatenating latent degradation map with skip-connection feature maps  (denoted as ``Our w/ concat injection").
For fair comparisons of computational cost, we add two-layer residual blocks after the feature concatenation. The operation causes a drop of 0.23 PSNR. We also remove the integration totally and concatenate the upsampling degradation maps with the blurry images at the network entrance with the blurry image (denoted as ``Ours input w/ concat''), which is the mainstream design in image denoising \cite{zhang2018ffdnet,MildenhallKPN18,Guo2019Cbdnet}.
Its performance drops by about 0.3 PSNR. All the ablation studies demonstrate the effectiveness of our proposed learnable degradation representations and MSDI-Net for image deblurring.
\section{Conclusions}
In this paper, we propose a framework for learning degradation representation with image reblurring and image deblurring. 
We first utilize an encoder to learn the degradation representations explicitly. Then we propose a multi-scale degradation injection network to effectively integrate the degradation representations for reblurring and deblurring.
With the degradation representations, our networks can be aware of and handle spatially varying degradation patterns adaptively. The experimental results demonstrate that our method outperforms other methods with a clear margin.

\clearpage
\bibliographystyle{splncs04}
\bibliography{egbib}

\begin{thebibliography}{10}
\providecommand{\url}[1]{\texttt{#1}}
\providecommand{\urlprefix}{URL }
\providecommand{\doi}[1]{https://doi.org/#1}

\bibitem{total_variation}
Chan, T., Wong, C.K.: Total variation blind deconvolution. IEEE Transactions on
  Image Processing  \textbf{7}(3),  370--375 (1998)

\bibitem{reblur2deblur}
Chen, H., Gu, J., Gallo, O., Liu, M.Y., Veeraraghavan, A., Kautz, J.:
  Reblur2deblur: Deblurring videos via self-supervised learning. In: 2018 IEEE
  International Conference on Computational Photography (ICCP). pp.~1--9 (2018)

\bibitem{HINet}
Chen, L., Lu, X., Zhang, J., Chu, X., Chen, C.: Hinet: Half instance
  normalization network for image restoration. In: Proceedings of the IEEE/CVF
  Conference on Computer Vision and Pattern Recognition (CVPR) Workshops. pp.
  182--192 (June 2021)

\bibitem{MIMO_UNet}
Cho, S.J., Ji, S.W., Hong, J.P., Jung, S.W., Ko, S.J.: Rethinking
  coarse-to-fine approach in single image deblurring. In: Proceedings of the
  IEEE/CVF International Conference on Computer Vision (ICCV). pp. 4641--4650
  (October 2021)

\bibitem{removing_non_uniform_motion}
Cho, S., Matsushita, Y., Lee, S.: Removing non-uniform motion blur from images.
  In: 2007 IEEE 11th International Conference on Computer Vision. pp.~1--8
  (2007)

\bibitem{gao2019dynamic}
Gao, H., Tao, X., Shen, X., Jia, J.: Dynamic scene deblurring with parameter
  selective sharing and nested skip connections. In: Proceedings of the IEEE
  Conference on Computer Vision and Pattern Recognition. pp. 3848--3856 (2019)

\bibitem{GAN}
Goodfellow, I., Pouget-Abadie, J., Mirza, M., Xu, B., Warde-Farley, D., Ozair,
  S., Courville, A., Bengio, Y.: Generative adversarial nets. In: Ghahramani,
  Z., Welling, M., Cortes, C., Lawrence, N., Weinberger, K.Q. (eds.) Advances
  in Neural Information Processing Systems. vol.~27. Curran Associates, Inc.
  (2014)

\bibitem{Guo2019Cbdnet}
Guo, S., Yan, Z., Zhang, K., Zuo, W., Zhang, L.: Toward convolutional blind
  denoising of real photographs. 2019 IEEE Conference on Computer Vision and
  Pattern Recognition (CVPR)  (2019)

\bibitem{Johnson2016Perceptual}
Johnson, J., Alahi, A., Fei-Fei, L.: Perceptual losses for real-time style
  transfer and super-resolution. In: European Conference on Computer Vision
  (2016)

\bibitem{VAE}
Kingma, D.P., Welling, M.: {Auto-Encoding Variational Bayes}. In: 2nd
  International Conference on Learning Representations, {ICLR} 2014, Banff, AB,
  Canada, April 14-16, 2014, Conference Track Proceedings (2014)

\bibitem{hyper_laplacian}
Krishnan, D., Fergus, R.: Fast image deconvolution using hyper-laplacian
  priors. In: Bengio, Y., Schuurmans, D., Lafferty, J., Williams, C., Culotta,
  A. (eds.) Advances in Neural Information Processing Systems. vol.~22. Curran
  Associates, Inc. (2009)

\bibitem{kupyn2018deblurgan}
Kupyn, O., Budzan, V., Mykhailych, M., Mishkin, D., Matas, J.: Deblurgan: Blind
  motion deblurring using conditional adversarial networks. In: {CVPR}. pp.
  8183--8192. Computer Vision Foundation / {IEEE} Computer Society (2018)

\bibitem{kupyn2019deblurgan}
Kupyn, O., Martyniuk, T., Wu, J., Wang, Z.: Deblurgan-v2: Deblurring
  (orders-of-magnitude) faster and better. In: {ICCV}. pp. 8877--8886. {IEEE}
  (2019)

\bibitem{sparse_prior}
Levin, A., Weiss, Y., Durand, F., Freeman, W.T.: Understanding and evaluating
  blind deconvolution algorithms. In: 2009 IEEE Conference on Computer Vision
  and Pattern Recognition. pp. 1964--1971 (2009)

\bibitem{Li2022Efficient}
Li, D., Zhang, Y., Law, K.L., Wang, X., Qin, H., Li, H.: Efficient burst raw
  denoising with variance stabilization and multi-frequency denoising network.
  International Journal of Computer Vision  \textbf{130}(8),  2060--2080 (Aug
  2022)

\bibitem{liang2021mutual}
Liang, J., Sun, G., Zhang, K., Van~Gool, L., Timofte, R.: Mutual affine network
  for spatially variant kernel estimation in blind image super-resolution. In:
  IEEE International Conference on Computer Vision (2021)

\bibitem{lim2017geometric}
Lim, J.H., Ye, J.C.: Geometric gan (2017),
  \url{https://arxiv.org/abs/1705.02894}

\bibitem{blind_spectral}
Liu, G., Chang, S., Ma, Y.: Blind image deblurring using spectral properties of
  convolution operators. IEEE Transactions on Image Processing
  \textbf{23}(12),  5047--5056 (2014)

\bibitem{cosine_annealing}
Loshchilov, I., Hutter, F.: {SGDR:} stochastic gradient descent with warm
  restarts. In: 5th International Conference on Learning Representations,
  {ICLR} 2017, Toulon, France, April 24-26, 2017, Conference Track Proceedings
  (2017)

\bibitem{contextual}
Mechrez, R., Talmi, I., Zelnik-Manor, L.: The contextual loss for image
  transformation with non-aligned data. arXiv preprint arXiv:1803.02077  (2018)

\bibitem{MildenhallKPN18}
Mildenhall, B., Barron, J.T., Chen, J., Sharlet, D., Ng, R., Carroll, R.: Burst
  denoising with kernel prediction networks. In: {CVPR} (2018)

\bibitem{SNGAN}
Miyato, T., Kataoka, T., Koyama, M., Yoshida, Y.: Spectral normalization for
  generative adversarial networks. In: International Conference on Learning
  Representations (2018)

\bibitem{restore_spatially_variant}
Nagy, J.G., O'Leary, D.P.: Restoring images degraded by spatially variant blur.
  SIAM Journal on Scientific Computing  \textbf{19}(4),  1063--1082 (1998)

\bibitem{deblur-multi-scale}
Nah, S., Kim, T.H., Lee, K.M.: Deep multi-scale convolutional neural network
  for dynamic scene deblurring. In: The IEEE Conference on Computer Vision and
  Pattern Recognition (CVPR) (July 2017)

\bibitem{checkboardartifacts}
Odena, A., Dumoulin, V., Olah, C.: Deconvolution and checkerboard artifacts.
  Distill  (2016)

\bibitem{blind_dark_channel}
Pan, J., Sun, D., Pfister, H., Yang, M.H.: Blind image deblurring using dark
  channel prior. In: 2016 IEEE Conference on Computer Vision and Pattern
  Recognition (CVPR). pp. 1628--1636 (2016)

\bibitem{park2020multi}
Park, D., Kang, D.U., Kim, J., Chun, S.Y.: Multi-temporal recurrent neural
  networks for progressive non-uniform single image deblurring with incremental
  temporal training. In: {ECCV} {(6)}. Lecture Notes in Computer Science, vol.
  12351, pp. 327--343. Springer (2020)

\bibitem{park2019SPADE}
Park, T., Liu, M.Y., Wang, T.C., Zhu, J.Y.: Semantic image synthesis with
  spatially-adaptive normalization. In: Proceedings of the IEEE Conference on
  Computer Vision and Pattern Recognition (2019)

\bibitem{Blind_DIP}
Ren, D., Zhang, K., Wang, Q., Hu, Q., Zuo, W.: Neural blind deconvolution using
  deep priors. In: 2020 IEEE/CVF Conference on Computer Vision and Pattern
  Recognition (CVPR). pp. 3338--3347. IEEE Computer Society, Los Alamitos, CA,
  USA (jun 2020)

\bibitem{realblur}
Rim, J., Lee, H., Won, J., Cho, S.: Real-world blur dataset for learning and
  benchmarking deblurring algorithms. In: {ECCV} {(25)}. Lecture Notes in
  Computer Science, vol. 12370, pp. 184--201. Springer (2020)

\bibitem{U-Net}
Ronneberger, O., Fischer, P., Brox, T.: U-net: Convolutional networks for
  biomedical image segmentation. In: Navab, N., Hornegger, J., Wells, W.M.,
  Frangi, A.F. (eds.) Medical Image Computing and Computer-Assisted
  Intervention -- MICCAI 2015 (2015)

\bibitem{Learning_to_deblur}
Schuler, C.J., Hirsch, M., Harmeling, S., Sch{\"o}lkopf, B.: Learning to
  deblur. IEEE Transactions on Pattern Analysis and Machine Intelligence
  \textbf{38},  1439--1451 (2016)

\bibitem{rotational_motion_deblurring}
Shan, Q., Xiong, W., Jia, J.: Rotational motion deblurring of a rigid object
  from a single image. In: 2007 IEEE 11th International Conference on Computer
  Vision. pp.~1--8 (2007)

\bibitem{Son2021PVDNet}
Son, H., Lee, J., Lee, J., Cho, S., Lee, S.: Recurrent video deblurring with
  blur-invariant motion estimation and pixel volumes. ACM Transactions on
  Graphics (TOG)  \textbf{40}(5) (2021)

\bibitem{GAT}
Starck, J.L, Murtagh, F., Bijaoui, A.: Image processing and data analysis. In:
  Cambridge University Press, Cambridge (1998)

\bibitem{suin2020spatially}
Suin, M., Purohit, K., Rajagopalan, A.N.: Spatially-attentive
  patch-hierarchical network for adaptive motion deblurring. In: {CVPR}. pp.
  3603--3612. Computer Vision Foundation / {IEEE} (2020)

\bibitem{Learning_CNN_non_uniform}
Sun, J., Cao, W., Xu, Z., Ponce, J.: Learning a convolutional neural network
  for non-uniform motion blur removal. In: 2015 IEEE Conference on Computer
  Vision and Pattern Recognition (CVPR). pp. 769--777 (2015)

\bibitem{tao2018srndeblur}
Tao, X., Gao, H., Shen, X., Wang, J., Jia, J.: Scale-recurrent network for deep
  image deblurring. In: IEEE Conference on Computer Vision and Pattern
  Recognition (CVPR) (2018)

\bibitem{Exploring_blur-CVPR21}
Tran, P., Tran, A., Phung, Q., Hoai, M.: Explore image deblurring via encoded
  blur kernel space. In: Proceedings of the {IEEE} Conference on Computer
  Vision and Pattern Recognition (CVPR) (2021)

\bibitem{DIP_2018_CVPR}
Ulyanov, D., Vedaldi, A., Lempitsky, V.: Deep image prior. In: Proceedings of
  the IEEE Conference on Computer Vision and Pattern Recognition (CVPR) (June
  2018)

\bibitem{unsupervised_degradation}
Wang, L., Wang, Y., Dong, X., Xu, Q., Yang, J., An, W., Guo, Y.: Unsupervised
  degradation representation learning for blind super-resolution. In: CVPR
  (2021)

\bibitem{wang2018pix2pixHD}
Wang, T.C., Liu, M.Y., Zhu, J.Y., Tao, A., Kautz, J., Catanzaro, B.:
  High-resolution image synthesis and semantic manipulation with conditional
  gans. In: Proceedings of the IEEE Conference on Computer Vision and Pattern
  Recognition (2018)

\bibitem{PMRID}
Wang, Y., Huang, H., Xu, Q., Liu, J., Liu, Y., Wang, J.: Practical deep raw
  image denoising on mobile devices. In: European Conference on Computer Vision
  (ECCV). pp. 1--16 (2020)

\bibitem{non-uniform-deblurring}
Whyte, O., Sivic, J., Zisserman, A., Ponce, J.: Non-uniform deblurring for
  shaken images. In: 2010 IEEE Computer Society Conference on Computer Vision
  and Pattern Recognition. pp. 491--498 (2010)

\bibitem{wang2018sftgan}
Xintao~Wang, Ke~Yu, C.D., Loy, C.C.: Recovering realistic texture in image
  super-resolution by deep spatial feature transform. In: IEEE Conference on
  Computer Vision and Pattern Recognition (CVPR) (2018)

\bibitem{L0_norm}
Xu, L., Zheng, S., Jia, J.: Unnatural l0 sparse representation for natural
  image deblurring. In: 2013 IEEE Conference on Computer Vision and Pattern
  Recognition. pp. 1107--1114 (2013)

\bibitem{Zamir2021MPRNet}
Zamir, S.W., Arora, A., Khan, S., Hayat, M., Khan, F.S., Yang, M.H., Shao, L.:
  Multi-stage progressive image restoration. In: CVPR (2021)

\bibitem{SAGAN}
Zhang, H., Goodfellow, I., Metaxas, D., Odena, A.: Self-attention generative
  adversarial networks. In: Chaudhuri, K., Salakhutdinov, R. (eds.) Proceedings
  of the 36th International Conference on Machine Learning. Proceedings of
  Machine Learning Research, vol.~97, pp. 7354--7363. PMLR (09--15 Jun 2019)

\bibitem{deepstacked}
Zhang, H., Dai, Y., Li, H., Koniusz, P.: Deep stacked hierarchical multi-patch
  network for image deblurring. In: {CVPR}. pp. 5978--5986. Computer Vision
  Foundation / {IEEE} (2019)

\bibitem{spatially_variant_recurrent}
Zhang, J., Pan, J., Ren, J., Song, Y., Bao, L., Lau, R.W., Yang, M.H.: Dynamic
  scene deblurring using spatially variant recurrent neural networks. In: 2018
  IEEE/CVF Conference on Computer Vision and Pattern Recognition. pp.
  2521--2529 (2018)

\bibitem{zhang2018ffdnet}
Zhang, K., Zuo, W., Zhang, L.: Ffdnet: Toward a fast and flexible solution for
  {CNN} based image denoising. IEEE Transactions on Image Processing  (2018)

\bibitem{zhang2020deblurring}
Zhang, K., Luo, W., Zhong, Y., Ma, L., Stenger, B., Liu, W., Li, H.: Deblurring
  by realistic blurring. In: {CVPR}. pp. 2734--2743. Computer Vision Foundation
  / {IEEE} (2020)

\bibitem{zhou2019stfan}
Zhou, S., Zhang, J., Pan, J., Xie, H., Zuo, W., Ren, J.: Spatio-temporal filter
  adaptive network for video deblurring. In: Proceedings of the IEEE
  International Conference on Computer Vision (2019)

\end{thebibliography}
\end{document}